# RIGHT IDEALS OF A RING AND SUBLANGUAGES OF SCIENCE

Javier Arias Navarro

Ph.D. In General Linguistics and Spanish Language

**http://www.javierarias.info/**

**Abstract**

Among Zellig Harris's numerous contributions to linguistics his theory of the sublanguages of science probably ranks among the most underrated. However, not only has this theory led to some exhaustive and meaningful applications in the study of the grammar of immunology language and its changes over time, but it also illustrates the nature of mathematical relations between chunks or subsets of a grammar and the language as a whole. This becomes most clear when dealing with the connection between metalanguage and language, as well as when reflecting on operators.

This paper tries to justify the claim that the sublanguages of science stand in a particular algebraic relation to the rest of the language they are embedded in, namely, that of right ideals in a ring.

## §1. Preliminary Word

In recent work (Arias 2015)[1] a line of research has been outlined in which the basic tenets underpinning the algebraic treatment of language are explored. The claim was there made that the concept of *ideal* in a ring could account for the structure of so-called *sublanguages of science* in a very precise way. The present text is based on that work, by exploring in some detail the consequences of such statement.

## §2. Introduction

Zellig Harris (1909-1992) contributions to the field of linguistics were manifold and in many respects of utmost significance. Nonetheless, not all of them achieved equal resonance in the field. Thus, the theory of what he labeled "science sublanguages" is widely overlooked and hardly understood, let alone appreciated. There are, of course, some noble exceptions, as we will see later (§5.2), but, for the most part, linguists and theorists of language have been oblivious to that core idea in Harris's methodology. This scenario becomes even more acute when leaving the English-speaking community. The topic has received only slight attention in French. My own endeavors in the Spanish-speaking world have panned out quite fruitless thus far. No work in this domain in German, Portuguese or Russian has come yet to my knowledge[2].

---

[1] Cf. Javier Arias, "Preámbulo a un análisis de la relevancia de los estudios de Ernst Kummer sobre factorización para la historia de las teorías del lenguaje", *Eikasia*, 64, May 2015, p. 53-80. In particular, on page 78 the reader may find an advance of the core of the present study:

"Sí puede resultar de interés, sin embargo, que le adelantemos al lector, de cara a futuros trabajos — por si estos llegaren a escribirse —, que al elaborar su teoría de los sublenguajes (que incluye, como caso egregio, los de la ciencia) Harris se apoya, precisamente, en el concepto de 'ideal' del álgebra abstracta. En concreto, define la relación de ciertos sublenguajes con el lenguaje en términos de un ideal a la derecha respecto a un anillo."

[2] In the crucial contributions regarding computational linguistics — machine learnability, definite clause grammars, semantic unification and so on — by Fernando Pereira (see, for instance, Pereira and Warren, 1980, Dalrymple, Shieber and Pereira 1991, or Pereira 2000) Portuguese plays a minor role and no real trace of an analysis of Portuguese sublanguages is to be found.

As to German, Lothar Hoffmann's extensive work on sublanguages certainly has to be reckoned with (e.g., Hoffmann 1985). However, its nature is that of a theoretical *Auseinandersetzung* leading to applications in the standardization of specialized vocabularies or terminological nomenclatures, as well as in the sociolinguistics behind text typologies, all of which are domains quite remote from the stance adopted here.

The present paper provides justification for Harris's statement that the sublanguages of science stand in a particular algebraic relation with the language they are a subset of. Such relation may be defined as that of a right ideal in a ring.

## §3. Sublanguages of Science

As starting point for our inquiry it will be wise to consider the following caveat:

"[…] the sublanguage grammar contains rules which the language violates and the language grammar contains rules which the sublanguage never meets. It follows that while the sentences of such science object-languages are included in the language as a whole, the grammar of these sublanguages intersects (rather than is included in) the grammar of the language as a whole."[3]

To put it graphically:

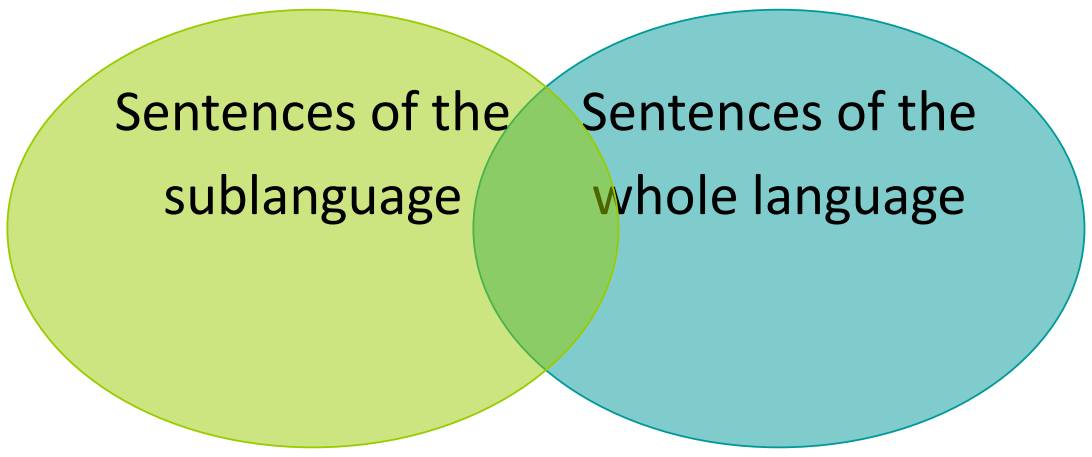

---

The term "sublanguage" should not be mistaken with "language for special purposes" (*LSP*). Unfortunately, Hoffmann sometimes seems to use both *Fachsprache* and *Subsprache* as synonyms.

[3] Cf. Zellig Harris, *Mathematical Structures of Language*, Robert E. Krieger Publishing Company, Huntington, New York, 1979, p. 154-155 [1st edition from 1968 by John Wiley & Sons, Inc.].

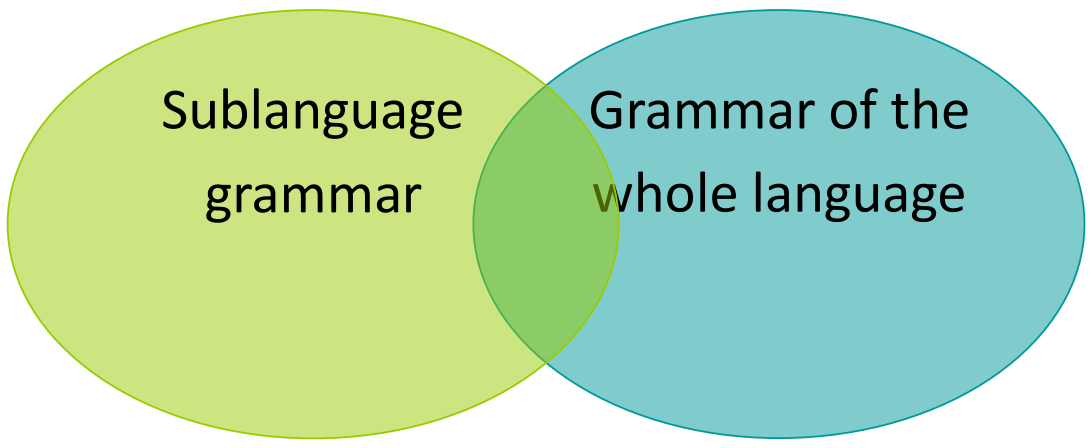

Let us briefly consider some cases. For instance, in the general language (English) a sentence such as *John activated protein A* would be well-formed; however, in the specific subject-matter domain (i.e. in the biomolecular sublanguage) the sentence is by no means legitimate, since the operator "*activate*" tightly constraints the combinations of the word classes at stake. Hence, the feature [+animate] (or strictly speaking, [person]) is discarded for the subject of a transitive sentence with the above-mentioned verb and a substance as syntactic object — and, accordingly, as semantic undergoer. Yet the sublanguage grammar allows for a substance to activate another substance or for a process to activate a substance. It is to be noted that none of those three options are excluded *a priori* in the grammar as a whole (that is, as a superset of the sublanguage in question).

The sublanguage operators exclusively reflect the salient relations and arguments that are meaningful and relevant in that given domain.

Harris correctly pointed out — as echoed by Stephen Johnson in his 1989 review[4] — that the sentence "The polypeptides were washed in hydrochloric acid" would be acceptable for a paper in a biochemistry journal, whereas "Hydrochloric acid was washed in polypeptides" would not. The reason is quite obvious: the subset is closed under syntactic operations of the general language. This implies that if a sentence is acceptable in the sublanguage, its syntactic transforms will be as well (as seen in "The enzyme activated the process.” → “The process was activated by the enzyme.”). Besides, within the sublanguage no sentences of the form Noun–Verb–Noun or alike are present; rather, a set of operators such as *is washed in* would have ordered-pair of arguments consisting, on the one hand, of a set of words for, say, molecules, cells and tissue and, on the other, of a set of words for acids, water and so on. Similarly, other operator subsets will yield different argument subsets.

[4] Cf. Stephen B. Johnson, "Review of *The form of information in science: analysis of an immunology sublanguage* by Zellig Harris, Michael Gottfried, Thomas Ryckman, Paul Mattick, Anne Daladier, T. N. Harris, and S. Harris", in *Computational Linguistics*, Volume 15, Issue 3, September 1989, p. 190-192.

Furthermore, whereas the whole English grammar theory basically deals with well-formed syntactic structures (or well-formedness constraints or conditions, if one prefers to put it that way), Harris’s sublanguage theory also includes domain-specific semantic information and relationships therein. The result is a language more informative than English, underscoring not only the syntactic structure but also the subject matter relations of a given domain.

Harris makes an important claim: when word combinations in a language are described in the most efficient way, a strong correlation is found between differences in structure and differences in information. Such correlation is even stronger in the sublanguages, where a more accurate correspondence between word combinations and information comes to the fore. The formal or quasi-formal systems built up by the sublanguages are consequently of great relevance when trying to characterize information (and the packing thereof) in the given sciences.

## §4. Ideal of a Ring

There are at least five different senses in which the term *ideal* can be used in mathematics, each of which can be identified with a given subdomain. To begin with, we have ideals in (algebraic) number theory. as developed by Richard Dedekind (1831-1916) out of Ernst Kummer’s (1810-1893) *ideal complex numbers*. Secondly, they may be found in abstract algebra, most notably in Emmy Noether’s (1882-1935) school. Furthermore, they play an important role in order theory, as can be seen in Marshall Harvey Stone (1903-1989) who, in virtue of the theorem stating the isomorphism from Boolean algebras to topological spaces that bears his name, may also be viewed as responsible for the set-theoretic sense of the term. Finally, a meaning of ideal can also be encountered in probability theory, tightly linked to sigma-ideals (σ-ideals)[5] and measure theory. Playing around a little bit with, say, the first three meanings (just for the sake of simplicity), one might jokingly end up with the figure of a parallel exact sequence as follows:

Number Theory → Abstract Algebra (Ring Theory) → Order Theory
↓ ↓ ↓
Ernst Kummer → Richard Dedekind, Emmy Noether → Marshall H. Stone

My concern here will be limited to the second meaning of them all, namely, the algebraic one, since it is the one Harris's theory hinges on. I am not specifically aware of when or by whom ideals of a ring were first introduced in (abstract) algebra, but Emmy Noether (1882-1935) did certainly deal with them in her 1929 paper "Hyperkomplexe Größen und Darstellungstheorie"[6]. In it, one may find the following definitions:

---

[5] A σ-ideal is an ideal (with additional conditions) in the representation of the σ-algebra Σ as a Boolean ring. According to the equivalences set up by Category Theory, every Boolean algebra can become a Boolean ring by taking the ring addition to be $A \oplus B = (A \sqcap \bar{B}) \sqcup (\bar{A} \sqcap B)$ and the ring multiplication to be $A \otimes B = A \sqcap B$. In the case of a subalgebra of the subset algebra, such as a σ-algebra, the ring operations in the Boolean ring are those of symmetric difference (with $\emptyset$ as additive identity) and intersection (with X itself as multiplicative identity). A ring ideal must be closed under multiplication (that is, intersection) with arbitrary σ-algebra elements, which amounts precisely to being closed under taking subsets. A ring ideal also has to be closed under finite unions, since $A \cup B = A \oplus B \oplus (A \cap B)$. Besides, a σ-ideal is closed under *countable* unions, which cannot be expressed in purely ring-theoretic language. I am deeply grateful to Henning Makholm for pointing out this to me. Needless to say, any errors or misconceptions are my sole responsibility.

[6] Cf. Emmy Noether, "Hyperkomplexe Größen und Darstellungstheorie", *Mathematische Zeitschrift*, 30, 1929, p. 641-692. Noether began her work on ring theory — as part of her whole enterprise pertaining to abstract algebra — as soon as 1920, in joint research with Werner Schmeidler (1890-1969), where a definition of left and right ideals in a ring was first provided (Cf. Emmy Noether and Werner Schmeidler, "Moduln in nichtkommutativen Bereichen, insbesondere aus Differential- und Differenzenausdrücken", *Mathematische Zeitschrift 8*: 1-2, 1920, p. 1-35. In particular, on page 7 the contrast between right-sided modul (*rechtsseitige Modul*) and left-sided modul (*linksseitige Modul*) is established and applied to polynomials. Note 10 on the same page pays tribute to Adolf Hurwitz as a forerunner, for his pioneering work with one-sided ideals when dealing with quaternions. Yet the very nature of his research made him automatically restrict his attention to principal ideals. The reader may get an update on these topics by consulting Anton S. Mosunov, "Ideal Class Group Algorithms in the Ring of Integral Quaternions", in http://arxiv.org/abs/1311.3379, 2014.

Noether's paper from the year after, *Idealtheorie in Ringbereichen* is generally reckoned as the foundation of general commutative ring theory. In it, Noether came to the very important result that in a ring which satisfies the ascending chain condition on ideals, every ideal is finitely generated.

A thorough historical discussion on the topic (undoubtedly, a far-reaching proposition for linguists to come) not only goes beyond the scope of this paper, but also wildly surpasses my best abilities. A very "*G* sei ein Ring, d.h. eine Abelsche Gruppe gegenüber Addition, wo auch eine Multiplikation definiert ist, mit den Eigenschaften

r(a+b)(a+b)rab.c = ra+rb = ar+br = a.bc

Jedes Element r definiert zugleich zwei Operatoren: die Operatoren rx und xr. Zugelassene Untergruppen sind die „Ideale", und zwar:

*linksseitige*, die die Operationen rx gestatten: ra⫅a
*rechtsseitige*, die die Operationen xr gestatten: ar⫅a
*zweiseitige*, die beide Operationen gestatten." [7]

Noether's original terminology can still be occasionally found in the literature in German. However, the terms *Linksideal* and *Rechtsideal* are more common today.
It should also be underlined that if *R* is any ring and *I* any left ideal in *R*, then *I* constitutes a left module over *R*. Analogously, right ideals are right modules.
In order to properly grasp the concept of one-sided module (and ideal), it is advisable (if not necessary) to first define *binary operation*[8] and their types. Let M and N be sets and M × M and M × N Cartesian products. An *internal binary operation* is to be construed as a function of the form M ×M → M, that is, as an operation which takes two objects from one single set and returns an object of the same type. By contrast, an *external binary operation* takes an object of type A and an object of type B and returns an object of type A. It follows, in that case, that A is closed with regard to the function at stake (an alternative phrasing would read: the internal operation requires the function to operate within the given set M). An external operation on M, on its turn, consists of a function M × N → M. In other words, an external binary operation is nothing else but a binary function from *K* × *S* to *S*, where *K* stands for a field (from the German word *Körper*) and *S* represents a vector space over that field. Contrary to a binary operation in the strict sense, *K* need not be *S*; its elements, as the very label points out, come from *outside*.

---

insightful overview is to be found in Bartel Leendert van der Waerden *History of Algebra*: from al-Khwarizmi to Emmy Noether, Springer Verlag, New York, 1986.

[7] Cf. Emmy Noether, *op. cit*., p. 646.

[8] The following paragraphs, up to the end of the section, have greatly benefited from discussion with Alessandro Pezzoni as well as from some comments by Mariano Suárez Álvarez and Najib Idrissi. The usual disclaimers apply here.

Vector addition and scalar multiplication on a vector space constitute typical examples of internal and external operation, respectively[9], of which the module operations considered here are a direct generalization.

It is worth mentioning on passing that the above-mentioned operations go beyond the scope of narrow algebra. Thus, the cup product known in algebraic topology can be viewed as a binary operation of fundamental nature in the realm of cohomology. As a matter of fact, it allows for both the internal and external interpretation, depending on certain conditions:

1) External cup product
   $H^*(X) \otimes H^*(Y) \to H^*(X \times Y)$ (sometimes called the cross product),
2) Internal cup product
   $H^*(X) \otimes H^*(Y) \to H^*(X)$ (which is obtained from the former by taking Y=X and composing with the map induced by the diagonal inclusion $X \to X \times X$).

We can now provide a suitable definition of *right module* (*Rechtsmodul*). Let R be a ring and M an Abelian group. M together with an external operation $M \times R \to M$, defined by $(x, \alpha) \mapsto x \cdot \alpha$, is called a right module over R if and only if the following requirements are met:

($M'_1$) $x \cdot (\alpha + \beta) = x \cdot \alpha + x \cdot \beta$ (Distributive Law)

($M'_2$) $(x + y) \cdot \alpha = x \cdot \alpha + y \cdot \alpha$

($M'_3$) $x \cdot (\alpha\beta) = (x \cdot \alpha) \cdot \beta$ (Associative Law)

---

[9] Dot product does not fall in either category. Nor does it, for instance, the function which takes two lengths and multiplies them to yield a surface in Physics. Sometimes a split is proposed within the external operations to make room for cases like the ones just quoted, taking two objects from a very same set A and yielding an object of a completely different set C, which would be then referred to as external (binary) operations of a second type, as opposed to the first type represented by scalar multiplication. That line of argumentation will not be pursued here.

for all $\alpha$, $\beta \in R$ and $x$, $y \in M$. Additionally, in those cases in which R also has an identity element, the module is called unitary if for all $x \in M$ the Unitary Law $x \cdot 1 = x$. ($M'_4$) holds

It goes without saying that, analogously, a left module owes its name to the fact that the element $\alpha \in R$ takes its place on the left side of the binary function.
The difference between left and right modules is not merely formal. This can be shown via *reductio ad absurdum*[10].

One-sided ideals (and modules) can therefore be interpreted, as we have just seen, as binary external operations.

In a noncommutative scenario, it becomes imperative to distinguish the *right ascending* chain condition on principal ideals (usually abbreviated to *ACCP*) from its *left* counterpart. The former just requires the poset of ideals of the form *xR* to satisfy the ascending chain condition, whereas the latter only computes the poset of ideals of the form *Rx*.

**§5. A mathematical illustration: matrices**

Before focusing on the linguistic material, it will be convenient to briefly consider how the newly-introduced concepts come to a particular instantiation in certain mathematical objects. Matrix calculus will provide an excellent example of that.

As the reader will already know, matrices are one of the most prominent cases of a non-commutative object under the product operator. In other words, given the matrices A and B, it holds that $AB \neq BA$. Thus, the basic prerequisite for the inquiry into one-sided ideals, namely, the condition of non-commutativity, is met.

---

[10] The specific case dealt with a little later in section §5 constitutes a perfect counterexample against the claim that a left module can be turned into a right module (and vice versa) under any circumstances. Note that such claim is also not automatically substantiated every time R is not commutative, since there are non-commutative rings in which every left ideal is also a right ideal (e.g. the ring of Hamilton's quaternions).

Speaking in a more formal way: let R be a noncommutative ring. A right ideal is a subset *I* which is an additive subgroup of R, such that for all $r \in$ R and all $a \in I$,

$a\,r \in I.$

For all $a \in$ R, the set

$$\langle a \rangle = \{r \in R \mid a\,r\}$$

is a right ideal of R, called the right ideal generated by *a*.

A right ideal need not be a two-sided ideal. The most obvious example of one-sided right ideals is the subset of quadratic matrices of order 2 which are zero on the bottom row. In other words, in the ring R of 2x2 matrices with entries in R, the subset

$$I = \left\{ \begin{bmatrix} a & b \\ 0 & 0 \end{bmatrix} \middle| \, a, b \in \mathbb{R} \right\}$$

is a right ideal. It is quite obvious that we are dealing with an additive subgroup here. Let us now briefly check whether or not the multiplication property holds:

$$\begin{bmatrix} a & b \\ 0 & 0 \end{bmatrix}\begin{bmatrix} c & d \\ e & f \end{bmatrix} = \begin{bmatrix} a\,c + b\,e & a\,d + b\,f \\ 0 & 0 \end{bmatrix} \in I.$$

For that *I* to be a one-sided ideal, it cannot be at the same time a left ideal. This is easy to prove, given that

$$\begin{bmatrix} 1 & 0 \\ 0 & 0 \end{bmatrix} \in I,$$

but

$$\begin{bmatrix} 0 & 0 \\ 1 & 0 \end{bmatrix}\begin{bmatrix} 1 & 0 \\ 0 & 0 \end{bmatrix} = \begin{bmatrix} 0 & 0 \\ 1 & 0 \end{bmatrix} \notin I.$$

Therefore we have a one-sided ideal which is a right ideal.

Summarizing, a right ideal of R is just a submodule of the right R module $R_R$ using the operation in R. To put it in the conventional symbols: a right ideal T of a ring (R,+,·) is a subgroup of (R,+) such that for all t ∈ T, r ∈ R, we have t·r ∈ T. The quotient $R_R/T_R$ is itself only a right module: it is a ring only if T is a two-sided ideal.

A different and yet interesting issue concerns the intuition behind the invention of ideals. An overview of the role they played with regard to the problem of unique factorization of primes is provided in Arias (2015). However, there is more than that to their genesis. Thus, this usage of the term *ideal* does not coincide with other sense often encountered in algebra according to which it refers to kernels of homomorphisms.

## §5.1 Some examples

After having just presented a case study in detail regarding a right ideal on a ring in mathematics (more specifically, on matrix rings) it is now time to turn to the linguistic instantiations of such phenomenon.

To begin with, reference is to be made to the ensuing paragraph, which encapsulates Zellig Harris's approach to the topic:

"Because a conjunction requires that the sentence following it have certain similarities to the sentence preceding (or, that particular kinds of sentences be further conjoined to compensate for the excess dissimilarities), each *SCS…CS* preserves certain properties of its initial *S*. As a result, certain sublanguages, including the science languages, have, under $C_S$, a relation to the language as a whole (in respect to all other operations on *S*) similar to that of a right ideal in a ring. If $S_1$ is in the sublanguage, and $S_2$ is not, $S_1$ $C_s$ $S_2$ retains properties of $S_1$ and is in the sublanguage. But in some cases, this holds only for those conjunctions which require strong similarities; or else it holds if the special grammatical properties of the sublanguage are defined only on the first *S* of each of its *SCS…CS*.

The import of this right-ideal type of construction is that certain subject-matter restrictions (referring thus to meaning, as against the material-implication conjoinings of sentences) are determined for every *SCS...CS* sequence by its first sentence (in the time order in the underlying regular form, i.e., before permutations). The possibility arises of covering the language with a system of such right-ideal-like subsets, whose intersection may be empty or may consist of certain distinguished sublanguages." [11]

Harris illustrates his theory in several places. An initial presentation of it can be found in his 1968 work. A concise summary is provided in the second chapter of

[11] Cf. Zellig Harris, *op.cit*., p. 155.

Harris's 1988 book[12]. A long detailed analysis constitutes the major enterprise of the monumental *The form of information in science*, in collaboration with Michael Gottfried, Thomas Ryckman, Paul Mattick Jr. and Anne Daladier, as well as with his brother Tzvee N. and sister-in law Suzanna Harris, Tzvee's wife, both professional immunologists[13].

I will stick to the brief version here, since it already contains the essential points which were to be empirically extended later on by Harris and his associates. Let us then turn to some of the main findings of such research. They may be listed as follows:

(I) A hypothesis is to be tested, namely, that changes in the grammar of a certain sublanguage over time reflect conceptual modifications in the field, and viceversa.

(II) The hypothesis is empirically tested on a subfield of biology, namely, immunology[14]. In particular, a collection of scientific papers from the period between 1935-1966 is searched. At that particular time-span, the main research question in the field was to determine which kind of cell was the producer of antibodies. More specifically, there was a heated controversy back then as to whether limphocytes or plasma cells triggered antibody production.

(III) An analysis is carried out in which similar combinability (of words and phrases) and patterns of syntactic occurrence are taken as core criteria. On those grounds, fifteen different classes were found. A formulaic notation is proposed which encompasses those classes and their interaction through several (verbally related) processes.

(IV) The conclusion is drawn that the empirical discoveries regarding antibody production automatically entail (and presuppose) a shift from some type of constructions to others (e.g. from AVT to AVC), which constitutes evidence supporting the initial assumption on conceptual and sublanguage change. This will be illustrated in some more detail in the paragraphs below.

[12] Cf. Zellig Harris, *Language and Information*, Bampton Lectures in America, 28, New York, Columbia University Press, 1988.

[13] Cf. Zellig Harris, Michael Gottfried, Thomas Ryckman, Paul Mattick Jr., Anne Daladier, Tzvee N Harris and Suzanna Harris, *The Form of Information in Science: Analysis of an Immunology* Sublanguage, Boston Studies in the Philosophy and History of Science, Dordrecht / Boston, Kluwer Academic Publishers, 1988. Anne Daladier carried out the analysis for French. The volume includes a preface by Hilary Putnam.

[14] The reader will find a useful overview on the topic in Carol Friedman, Pauline Kra and Andrey Rzbetsky, "Two biomedical sublanguages: a description based on the theories of Zellig Harris", *Journal of Biomedical Informatics*, 35, 4, 2002, p. 222–235.

According to Grishman and Kittredge's (1986) definition, a sublanguage is "a subsystem of language that behaves essentially like the whole language, while being limited in reference to a specific subject domain"[15]. As they correctly point out, "each sublanguage has a distinctive grammar, which can profitably be described and used to solve specific language-processing problems"[16]. Such grammar exhibits a specialized vocabulary, idiosincratic semantic relationships, and sometimes (see §.3) specialized syntax as well. Forms of language such as those used in weather reports, scientific articles or real estate advertisements count among the most notorious and usually quoted samples of sublanguage.

The main assumption of the theory of sublanguages of sciences is then somewhat reminiscent of the Bloomfieldian statement according to which "the division of labor, and with it, the whole working of human society, is due to language"[17].

---

[15] Cf. Ralph Grishman and Richard Kittredge, Preface to Ralph Grishman and Richard Kittredge (eds.). *Analyzing Language in Restricted Domains: Sublanguage Description and Processing.* Lawrence Erlbaum Associates, Hillsdale, New Jersey, 1986, p. ix. There is a recent reprint by Psychology Press, New York / London, 2014.

A fairly detailed historical overview on the development of the term is provided in Sharon O'Brien, *Sublanguage, Text Type and Machine Translation*, Master Thesis at Dublin City University, Dublin, 1993, p. 2–14.

[16] Cf. Ralph Grishman and Richard Kittredge, *op. cit.*, p. ix. With regard to the implementation of language-processing techniques, see Ralph Grishman, "Adaptive Information Extraction and Sublanguage Analysis", *Working Notes of the Workshop on Adaptive Text Extraction and Mining, Seventeenth International Joint Conference on Artificial Intelligence (IJCAI-2001),* Seattle, Washington, August 5, 2001, as well as Ralph Grishman, Lynette Hirschman, Ngo Thanh Nhan, "Discovery Procedures for Sublanguage Selectional Patterns: Initial Experiments", *Computational Linguistics*, Volume 12, Number 3, July-September 1986, p.205-215. Research carried out in Montréal in the late 60's and 70's by Alain Colmerauer and collaborators (like Richard Kittredge himself) yielded the first prototype of a sublanguage-based machine translation system, TAUM METEO, capable of translating weather reports from English into French with high accuracy.

[17] Cf. Leonard Bloomfield, *Language*, New York, Holt, Rinehart and Winston, 1933, p. 24. The epistemology of the great American linguist, often associated with the Vienna Circle, came to be expressed in his remarkable paper "A set of postulates for the science of language", *Language*, 2 (3), 1926, p. 153–164. Recently, interest on Bloomfield's theory of science has experienced some revival, as attested, for instance, by Thomas Meier's conference, "A logical Reconstruction of Leonard Bloomfield's Linguistic Theory", given on 30th March 2012 at the Munich Center for Mathematical Philosophy (an online preprint of the resulting paper can be found in http://philsci-archive.pitt.edu/9405/).

A set of argument classes and operators is established. The former includes categories such as *antibody* (A), *antigen* (G), *cell* (C), *tissue* (T), and *body part* (B); the latter makes reference to processes of the kind referred to by *inject* (J), *move* (U), or *present in* (V).

The sublanguage sentences can be described then as synthetic formulae (or derived from such, depending on your view). Some notorious examples quoted in the literature are listed below:

G J B "antigen was injected into the foot-pads of rabbits"
A V C "antibody is found in lymphocytes"
G U T "antigen arrives by the lymph stream"

A V C "antibody appears in plasma cells"
A V C "antibody is found in plasma cells"[18]

Some further notational conventions apply: individual formulae with subscripts for the different class members are used, while superscripts stand for the corresponding modifiers.

The complex linguistic system so obtained presents precisely the information of the given science, with no loss from the original presentation in a natural language.

A basic tenet of the theory of sublanguages is that changes in information over time in a given science and / or subject matter can be located and characterized in the sentence structures. This amounts *de facto* to a conceptual genealogy or, in other words, to a diachrony which accounts for the emergence of new views in a scientific field by resorting to observed disagreements between (at least) two different information-carrying sentence-forms in time. For instance, a shift in time is observed from AVT to AVC, corresponding to the better understanding of the nature of cell types provided by more precise electronic microscopes, as opposed to the previous stage in which the different cell types were not easily distinguished in the tissue. Later on, a new type appears, namely CYC, once more kinds of cells came to be distinguished, and their similarities started to be pointed out. Finally, formulae with the CCYC pattern occurred more and more often, to the brink of becoming the prevailing pattern, along with the spreading view that a cell is a later stage of a previous cell.

[18] It should not be overlooked that in the grammar of the whole language, *is found* ought to be analyzed as a passive voice transformation from *Someone finds antibody in plasma cells*.

The diachronic sequence can be represented as follows:

AVT > AVC > CYC > CCYC

This drift corresponds to an increasing differentiation in cell types, whose number is in turn controlled by the claim that some different names indeed identify the very same cell class.

It has been seen thus far that the term *ideal* is in fact polysemic and can occasionally lead to misunderstandings. One should at least differentiate two meanings:

1) Ideals as kernels of ring homomorphisms. The analogue of ideals for groups would be normal subgroups.

2) In number theory, ideals appear naturally as the appropriate objects for the study of *divisibility.* The statement that *a* divides *b* is nothing more than the statement that *b* is contained in the ideal (*a*) generated by *b*, or, using the language of ring homomorphisms, that $b = 0$ in the quotient ring R/(*a*). Therefore, instead of studying divisibility of numbers one can just as well study containment of ideals. For Z this yields the same theory of divisibility; however, for other number rings that results in non-principal ideals, which are precisely the manifestation of Kummer's "ideal numbers".

The possibility of a spatial (geometrical or topological) representation for ideals and, subsequently, for any object or domain they might underlie, is warranted by the Riemann-Roch Theorem[19], among others.

---

[19] Cf. Martin Krieger, *Doing Mathematics: Convention, Subject, Calculation, Analogy*, New Jersey, World Scientific Publishing, 2003, p. 223.
"Hilbert then shows how one of Dedekind's notions of a prime factor or ideal (the different) corresponds to the Riemann-Roch theorem, a geometric and arithmetic fact concerning the topology of Riemann's surfaces".
Dedekind's concept of what later came to be known as *different ideal* or *the different* can be found for the first time in his 1882 paper "Über die Discriminanten endlicher Körper", *Abhandlungen der Königlichen Gesellschaft der Wissenschaften zu Göttingen*, 29, 2, p. 38, under the name *das Grundideal des Körpers* being denoted by $\theta^*$. It stands, as Dedekind himself points out, in a tight relation to the *discriminant*

Once some fundamental concepts have been presented, a couple of questions come to the fore right away:

1) Is there the same number of right and left ideals?

2) Which is the reason for Harris choosing right ideals instead of left ideals for his description of the sublanguages of science?

Trying to provide an answer to 1) and 2) becomes crucial for the goal of the present paper. With regard to 1), the theoretical possibility should first be entertained of every non-commutative ring always having exactly the same number of left ideals as that of right ideals, while assuming that both types of ideals should present equal length[20]. However, empirical evidence soon makes us discard this hypothesis[21]. For instance, consider a field endomorphism σ: F→F such that [F: σ (F)] = n > 1. Now take the Ore extension[22] F [$x$; $\sigma$] and let $R$ = F [$x$; $\sigma$] / ($x^2$). Let the image of $x$ in the quotient

---

("*Discriminante*" or "*Grundzahl*"). The rationale behind the different ideal is to carry out an analogy in number fields to the geometric notion of a dual lattice in Euclidean space.

[20] *Length* is to be taken here in the sense of 'length of a module'. Or, with a better formulation: the length of a ring R is to be interpreted as an R-module, which means that Rx is the only left ideal of R, so the maximal chain of left R modules contained in R is Rx⊂R which has length 2 (or rather 1, given the convention of counting chains from 0). Now, Rx counts as maximal but not minimal as a right R-module, so it follows that there is a chain of at least length 3 of right R-modules contained in R.

In other words, let R be a ring. The length of a left (or right) R-module M is the maximum length $n$ of a chain of left (or right) R-modules $M_0 \subset M_1 \subset \cdots \subset M_n = M$, if it exists; otherwise it is said to be infinite. The length of R is thus its length as a left (or right) R-module. Note that the only module of length 0 is the zero module, which is contained in every other module. Furthermore, it can be shown that a module (or ring) has finite length if and only if it is both Noetherian and Artinian. Needless to say, the lengths of a ring seen as a left or right module can differ, even if they are both finite.

The above definition entails that the notion of length is of limited use for commutative rings, given that a commutative Noetherian ring is Artinian if and only if all of its prime ideals are maximal. Specifically, no integral domain can have finite length, because its 0 ideal is prime. For instance, if $n \in$ Z is different from 0 or ±1, then Z ⊃ ($n$) ⊃ ($n^2$) ⊃ ($n^3$) ⊃··· be denoted by $\bar{x}$. It can be verified that $Rx$ = F $\bar{x}$ is a minimal and maximal left ideal of R as well as a maximal right ideal. As a right R module it is semisimple, with composition length *n*. Thence the composition length of $_R R$ equals 2, whereas the composition length of $R_R$ is n+1 > 2.

[21] That result should come as no surprise, for we analogously know, as a matter of fact, that a non-commutative ring may have different numbers of left ideals and two-sided ideals. Thus, a matrix ring over a field has just 2 two-sided ideals, i.e. the trivial cases of the unity and zero, while presenting some non-trivial left ideals.

[22] An Ore extension of a skew field is, by definition, a non-commutative principal ideal domain (PID), where xa = σ(a)x, for all a ∈ F. The original development can be traced back to Øystein Ore's seminal paper "Theory of non-commutative polynomials", *The Annals of Mathematics*, 34, 1933, pages 480-508. An overview of the topic in noncommutative algebras is provided, for instance, by Ken R. Goodearl and Robert B. Warfield Jr., *An introduction to non-commutative Noetherian rings*, London Mathematical Society Student Texts, Vol. 16, Cambridge, Cambridge University Press, 1989, or by Paul Moritz Cohn in *Skew Fields: Theory of General Division Rings*. Cambridge, London, 1995.

Once some counterexample regarding length has been provided, let us pay attention to the number of one-sided ideals *per se*: it can be shown that, given a field endomorphism $\rho$: F→F, such that [F: $\rho$ (F)] ≥ 2, it yields a ring with three left ideals, whereas the number of right ideals will be never below four. Therefore, the number of right ideals does not need to coincide with that of left ideals.

Another valid example in this regard is made up by the subring *R* of the matrix ring $M_2$(R) of all matrices of the form

$$\begin{pmatrix} a & b \\ 0 & c \end{pmatrix}$$

where $a \in$ Q, and $b, c \in$ R.

Let us suppose we find a phenomenon (in nature, social organization, whatever) for which we believe (as some author may have stated) a formalization or modeling in terms of ideal of a ring to be possible. Let us further suppose the possibility of a two-sided ideal is excluded (what specific feature of the phenomenon might exclude that?). The question which immediately comes to mind is: which kind of property of the phenomenon or object of study would be responsible for the decision between a right or a left ideal of the ring? A possible (albeit somewhat cynical) answer would be to say there is no real reason to decide, since a left ideal of a ring is the same as a right ideal of the opposite ring. It should be borne in mind that every ring has an opposite[23]: thus, for the case of the ring of nxn matrices (more specifically, of the 2x2 matrix type) one comes to the following result: the matrix ring $M_n(K)$ is isomorphic to its opposite ring via $A \mapsto A^T$, in so far as K is commutative, or has an involution, or fulfills some other sufficient condition.

[23] There is always, indeed, an opposite ring available. Even so, it just might not be one that can be described “naturally”. For the meaning of the concept ‘naturality’ in mathematics, the reader can find some useful remarks in Martin H. Krieger, *op. cit.*, p. 89-90.

Another line of argumentation goes as follows: the answer just depends on how one sets up the algebraic model. If actions are written from left to right, most likely work will be done with left ideals. Conversely, if they are written form right to left, right ideals will come into play. This would be analogous to the choices people make on writing functions. For that to hold, that is, for the issue to be reduced to a mirage of the notational system adopted, the assumption should be made that all phenomena at stake are symmetric.

Could it be rather that a phenomenon should be formalized as a right ideal, with no opposite ring available at the natural or social description at stake? Given that every nonzero ring has a maximal right ideal, and since a right ideal of any ring automatically is a left ideal of its opposite ring, it follows that a maximal right ideal of a nonzero ring is a maximal left ideal of its opposite. It seems, then, that only practical reasons or even personal preferences might have made Zellig Harris choose right instead of left ideals as the basis for his reflection on sublanguages.

Let us now briefly consider some mathematical principles which will help shed some further light on the topic. To begin with, let us refer to an important result in algebra from back in 1950:

Cohen's *Theorem*: A commutative ring $R$ is Noetherian iff every prime ideal of $R$ is finitely generated[24].

The above theorem, together with another stating that a commutative ring $R \neq 0$ is a Dedekind domain iff every nonzero prime ideal of $R$ is invertible, constitute the core, so to speak, driving the structure of the given ring. Now, assuming all this, Manuel Reyes crucially notes:

---

[24] Cf. Irvin Sol Cohen "Commutative rings with restricted minimum condition", *Duke Mathematical Journal*, 17 1950, p. 27–42. An overview of this and the ensuing topics in the remaining of this section is provided by David Eisenbud, *Commutative Algebra: with a View Toward Algebraic Geometry*, Graduate Texts in Mathematics, vol. 150, New York, Springer, 1995.

"While prime two-sided ideals are studied in noncommutative rings, it is safe to say that they do not control the structure of noncommutative rings in the sense of the two theorems above. Part of the trouble is that many complicated rings have few two-sided ideals." [25]

Language and sublanguages, one might argue, could easily fall under those complicated rings just mentioned, their structure being controlled by their one-sided ideals.

The concept of *Oka family* (after the great Japanese mathematician Kiyoshi Oka)[26] may prove useful at this point. It can be defined as follows: an ideal family $\mathcal{F}$ in a ring $R$ with $R \in \mathcal{F}$ is said to be an Oka family if, for $a \in R$ and $I, A \lhd R$, $(I, a)$, $(I : a) \in \mathcal{F} \Rightarrow I \in \mathcal{F}$. Accordingly, an ideal family $\mathcal{F}$ in a ring $R$ with $R \in \mathcal{F}$ constitutes an *Ako family* if, for $a, b \in R$ and $I, B \lhd R$, $(I, a)$, $(\mathrm{I}, b) \in \mathcal{F} \Rightarrow (I, ab) \in \mathcal{F}$.

The family of principal ideals of a ring, for instance, qualifies as an Oka family. If $\mathcal{F}$ is an Oka family of ideals, it follows that any maximal element of the complement of $\mathcal{F}$ must be prime. The same holds for Ako families. This can be expressed through the *Prime Ideal Principle* (*PIP*), which claims: if $\mathcal{F}$ is an Oka family or an Ako family, then any ideal maximal with regard to not being in $\mathcal{F}$ is prime.[27]

It should be underlined that the notions of Oka and so-called Ako families (a mere acronym creation inverting Oka's name) may be thought of as generalizations of the more widespread concept of a monoidal filter. It is relatively easy to find families of ideals that are Oka but not Ako. The opposite is much harder.

---

[25] Cf. Manuel L. Reyes, "A one-sided prime ideal principle for noncommutative rings", 2010, available at http://arxiv.org/pdf/0903.5295.pdf. The quote can be found on page 1 of that document.

[26] The ensuing definitions of Oka and Ako families are taken from Tsit Yuen.Lam and Manuel L. Reyes, "A Prime Ideal Principle in commutative algebra**,** *Journal of Algebra*, 319, 7, 2008, p. 3009.

[27] Cf. Tsit Yuen Lam and Manuel L. Reyes, "Oka and Ako Ideal Families in Commutative Rings", in *Contemporary Mathematics*, 480: *Rings, Modules and Representations*, American Mathematical Society, Providence, 2009, p.264.

An *Oka family of right ideals* (also known as *right Oka family*) in $R$ is a family $\mathcal{F}$ of right ideals with $R \in \mathcal{F}$ such that, given any $I_R \subseteq R$ and $a \in R$,

$$I + aR,\ a^{-1} I \in \mathcal{F} \Rightarrow I \in \mathcal{F}.$$

Now, the *Completely Prime Ideal Principle* (*CPIP*) generalizes the Prime Ideal Principle to one-sided ideals of a non-commutative ring, by giving a formal expression to the idea that right ideals that are maximal tend to be completely prime. The *CPIP* is thus the main tool yielding the connection between completely prime right ideals and the (one-sided) structure of a ring.

As a result of the *CPIP*, a noncommutative generalization of Cohen's Theorem can be provided.

Out of all of the above remarks new insight into the nature of sublanguages considered as right ideals might be gained, leading to a more subtle conceptual framework, as well as to the possibility of a more fine-grained representation and visualization of them. This is a task for linguists and mathematicians to come, who will certainly be more suited for it than the author of this paper.

The key for such eventual representation is provided by Anatoly Ivanovich Maltsev:

"[…] toda álgebra asociativa se puede sumergir en otra con elemento unidad. La representación regular de esta última es fiel; por consiguiente, esta representación del álgebra dada es también fiel. Así, pues, toda álgebra asociativa tiene una representación fiel por medio de matrices.

Este método de construir representaciones es insuficiente para hallar todas las representaciones de un álgebra. Existe otro método más fino que está relacionado con el concepto de «ideal». El concepto de ideal juega un importante papel en las matemáticas modernas.

Un sistema $I$ de elementos de un álgebra recibe el nombre de *ideal* a la derecha si es un subespacio lineal del álgebra y si el producto a la derecha de todo elemento de $I$ por cualquier elemento del álgebra es también un elemento de *I*. Un ideal a la izquierda se define de modo similar (intercambiando el orden de los factores). Un ideal a la derecha y a la izquierda simultáneamente se llama ideal bilátero. Es evidente que el elemento cero del álgebra forma por sí solo un ideal bilátero, que recibe el nombre de ideal cero del álgebra. El álgebra completa también es un ideal bilátero. Sin embargo, aparte de estos dos ideales, el álgebra puede contener otros, cuya existencia está habitualmente relacionada con interesantes propiedades del álgebra.

Supongamos que un álgebra asociativa *A* contiene un ideal a la derecha *I*. Tomemos en él una base $\epsilon_1, \epsilon_2, \ldots, \epsilon_m$. Puesto que en general *I* es sólo una parte de *A*, sus bases tendrán en general menos elementos que las de *A*. Sea $\alpha$ un elemento de *A*. Como *I* es un ideal a la derecha y $\epsilon_1, \epsilon_2, \ldots, \epsilon_m$ están contenidos en *I*, los productos $\epsilon_1\alpha, \ldots, \epsilon_m\alpha$ pertenecen también a *I* y se pueden expresar, por tanto, linealmente en función de la base $\epsilon_1, \ldots, \epsilon_m$; es decir,

$$\epsilon_1\alpha = a_{11}\,\epsilon_1 + \ldots + a_{1m}\,\epsilon_m$$

$$\ldots\ldots\ldots\ldots\ldots\ldots\ldots\ldots\ldots\ldots$$

$$\epsilon_m\alpha = a_{m1}\,\epsilon_1 + \ldots + a_{mm}\,\epsilon_m$$

Asociando a cada elemento $\alpha$ la matriz $\|a_{ij}\|$ obtenemos, como antes, una representación del álgebra *A*. El grado de esta representación es igual al número de elementos de la base del ideal y por tanto es, en general, menor que el de la representación regular. Evidentemente, la representación de menor grado obtenida por este método corresponderá a un ideal minimal. De aquí se deriva el fundamental papel de los ideales minimales en la teoría de álgebras.” [28]

---

[28] Cf. Anatoly Ivanovich Maltsev “Grupos y otros sistemas algebraicos”, in A.D. Aleksandrov, A.N. Kolmogorov, M.A. Laurentiev et al. (eds.), *La matemática: su contenido, métodos y significado*. Madrid, Alianza Universidad, 1994 (7th reprint.), volume III., §12, p. 392-393. For the sake of faithfulness, I am keeping the term “minimal” used by the Spanish translator instead of replacing it for the undoubtedly more appropriate “mínimo”.

Another definition of ‘ideal’, more focused on ring theory than on the theory of algebras, can be found in that very same volume, §14, p. 399-400.

A final remark should be made to close this section. It is an established fact that there are only three associative algebras with division on the field of real numbers, namely, the field of real numbers itself, the field of complex numbers and the algebra of quaternions. It has been noted recently (inasmuch as the parallelism with the study of language may be significant) that up to this day linguistics and the theory of language have only stepped into the soil of real elements and have not really trespassed the fences signaling the territories of complex units and quaternion-like entities[29].

I am fully convinced that taking that domain extension seriously would most certainly represent a quantum leap in the development of the discipline. Unfortunately, academic atavism has — once more — prevented this from happening.

### §5.2 Visual representation of one-sided ideals

The French linguist Gustave Guillaume, famous for his penchant for graphic representations of linguistic structures, used to justify his extensive use of diagrams by quoting Leibniz: "Things impede each other, ideas do not" [30]. Whichever the degree of agreement with his claim, it is undeniable that visualization may help intuitions crystalize in a clearer fashion, allowing for increasing naturality. It should be always kept in mind, however, that, as the saying goes, "a theorem is not true any more because one can draw a picture, it is true because it is functorial."[31]

In (§.3) a Venn diagram was provided which represents the inclusion relations between the sublanguage and the whole language grammar. However, in order to fulfill

---

[29] Cf. Arias 2015, *op. cit.*, p. 69, note 27.

[30] Cf. Gustave Guillaume, *Foundations for a Science of Language*, John Benjamins Publishning Company, Amsterdam / Philadelphia, 1984, p. 123-124:

"From his writings, Leibnitz appears to have felt this difference between what is perceivable or primary in the mind and what is expressible or secondary, and brought out in human language. That is why he so wisely advised thinking in diagrams. "Things impede each other, ideas do not". Diagrams are still things, but less so than the signs language uses to exteriorize its interiority. To think in diagrams is to keep things, to a large extent, from impeding each other. But to arrive at the exact diagram needed requires sustained reflection carried out with rigour and finesse. There is a real risk of constructing false diagrams which fortunately is greatly reduced by the fact that in order to construct a diagram one must start with extremely simple, elementary views with highly plausible exigencies."

[31] Cf. Serge Lang, "Review of A.Grothendiek and J. Dieudonné, *Éléments de géométrie algébrique*", *Bulletin of the American Mathematical Society* 67, 1961, p. 245.

the goal of an adequate and exhaustive description of the phenomenon, more precise ways are due. Thus, Hasse diagrams are of customary use when trying to visualize the structures known as *lattices*. Interestingly enough, in a book widely acknowledged among linguists, Partee, ter Meulen and Hall make use of *lattices* (together with the *meet* and *join* operations) in order to represent filters and ideals[32]. Such order theoretic notion of ideal has an exact equivalence in set theory, the relevant order being set inclusion.

It is worth mentioning that the important distinction between *prime ideals*, *semiprime ideals* and *primary ideals* finds a straightforward representation by means of a Hasse diagram, as can be seen in several handbooks on the matter.

In a previous section (§5) matrices — or, more specifically, a subset thereof — were shown to provide the best intuitive example of one-sided ideals in a ring. Hence, the question legitimately arises: to how to represent them graphically? It turns out that there is no one way to “visualize” matrices. As a matter of fact, mental images of abstract quantities, albeit certainly useful, often lead to loss of information.

If we limit our focus to 2x2 matrices, one way (but by no means the only) of grasping them graphically is to note that they have the same mathematical structure as complex numbers with regard to addition and multiplication[33] and therefore can be represented accordingly, by means of the usual Argand diagram.

A more useful approach is to view the matrix as a linear transformation on a vector and observe its action on a standard set of vectors. Linear transformations and their properties in terms of dilating and rotating vectors are useful ways of understanding matrices, although they are not the only relevant information included in a matrix[34].

---

[32] Cf. Barbara Partee, Alice ter Meulen and Robert E. Wall. *Mathematical Methods in Linguistics*, Dordrecht, Kluwer Academic Publishers, 1993, Chapter 11, “Lattices”. The discussion on filters and ideals comprises pages 285-287. One can generate a Hasse diagram for any finite poset. Needless to say, the inverse is also true: it is always possible to generate some lattice based on a given diagram. The most important trait regarding Hasse diagrams is that they minimize the number of edges used. In every Hasse diagram, the ordering of nodes that are not directly adjacent is stored implicitly by paths through the edges (hence the frequent term order-preserving mapping in this context).

[33] The proper study of hypercomplex numbers (to which matrices have so deeply contributed) began in 1870 when Benjamin Peirce first published in lithographic form his *Linear Associative Algebra* in Washington, and was carried forward by his son, the renowned philosopher Charles Sanders Peirce. For a more detailed account of these origins, see Ivor Grattan-Guinness “Benjamin Peirce's Linear Associative Algebra (1870): New Light on its Preparation and ‘Publication’”, *Annals of Science*, 54, 1997, p.597-606.

Getting a mental picture of a matrix is a truly hard goal. That difficulty may be explained as follows: suppose you have a door and you open it. You could prove that the door is in two different states because you did a certain transformation to it, namely, you actually rotated the door around the hinge. That was a transformation of a three-dimensional object in time. A transformation, in this view, is nothing but a structure in four-dimensional space-time for the door. One happens to be lucky here because only 2 coordinates, $x$, $y$ are indeed transformed (you cannot rise up the door when you open it).

A 2x2 matrix could actually be seen in four-dimensional space. But that would also imply that for you opening a door is actually meaningless in 4D space (not in time, though) because you see a geometrical object that has all the transitory states between the two states (door opened in state 1 or state 0). In time, this would mean that events are seen in space-time just as single stationary objects. For example, a man speaking to me would not actually be speaking to me. I would see him before he spoke and as he speaks and for me it all would constitute a single object or event, even though from the perspective of mere mortals he would have gone through an infinite number of transitions.

Another answer goes along the following lines: 2×2 matrices

$$A= \begin{pmatrix} a & b \\ c & d \end{pmatrix}$$

can be visualized in the form of a parallelogram $Q_A \subset \mathrm{R}^2$ with vertices $(0,0),(a,c),(b,d)$ and $(a+b,c+d)$. If one identifies the plane $\mathrm{R}^2$ with $\mathrm{M}^{2\times1}(\mathrm{R})$, the space of 2×1 column matrices, then $Q_A$ is the image of the unit square $[0,1]\times[0,1]$ under linear transform

$$[0,1] \times [0,1] \ni \begin{pmatrix} x \\ y \end{pmatrix} \mapsto \begin{pmatrix} x' \\ y' \end{pmatrix} = \begin{pmatrix} a & b \\ c & d \end{pmatrix} \begin{pmatrix} x \\ y \end{pmatrix} \in Q_A.$$

---

[34] Michael Artin's book *Algebra*, Lebanon, Indiana, Prentice-Hall., 1991, constitutes a very valuable source highlighting some of these connections.

Since a linear transformation is uniquely determined by its action on a basis, this faithfully represents the 2×2 matrices. Under this representation, some geometry-related operations now correspond to familiar geometric shapes. Thus, the matrix

$$\begin{pmatrix} s & 0 \\ 0 & s \end{pmatrix}$$

represents a scaling of geometric objects. It corresponds to a square of side length $s$, axis aligned with the standard $x$- and $y$-axis.

For instance, the matrix

$$\begin{pmatrix} \cos\theta & -\sin\theta \\ \sin\theta & \cos\theta \end{pmatrix}$$

represents a counterclockwise rotation of angle θ. It corresponds to the unit square rotated counterclockwisely for angle θ, assuming the polar form of representation.

The matrices

$$\begin{pmatrix} 1 & m \\ 0 & 1 \end{pmatrix}$$

and

$$\begin{pmatrix} 1 & 0 \\ m & 1 \end{pmatrix}$$

represent sheer mappings in horizontal and vertical directions. They can be visualized as a parallelogram with one pair of its sides staying horizontal and vertical respectively[35].

---

[35] This sort of shapes provides a useful visual mnemonics for what the effects of those matrices (when viewed as a transformation of the plane) are.

Last but not least, one might also use the same idea to introduce the concept of determinant. The determinant of a matrix $A$ is just the area of $Q_A$. One might then ask: what does det $A < 0$ mean? The answer is pretty straightforward: that $Q_A$ has been flipped.

Given the assumption that sublanguages stand to the whole language in a relation characterized as that of right ideals of a ring, it can be naturally concluded that they will share their visual representation. Even if that is logically true, the fact remains that the representation of sublanguages we come across in the literature is formulated with more basic means. Thus, Lehrberger (1986) provides a figure showing how any given sublanguage intersects both the standard language and the language as a whole[36].

In some ways, one might argue, representation of sublanguages has historically lain behind that of the ideals underlying them. Making up for this deficit should constitute a desirable (as well as reasonable) goal in the discipline.

## §5.3 Metalanguage and Language

A very important feature of Harris's view relates to his analysis of metalanguage. Following Kurt Gödel's path, he correctly points out that the metalanguage is contained in the language:

"Every natural language must contain its own metalanguage, i.e., the set of sentences which talk about any part of the language, including the whole grammar of the language. Otherwise, one could not speak in a language about that language itself; this would conflict with the observation that in any language one can speak about any subject, including the language and its sentences, provided that required terms are added to the vocabulary. Furthermore, there would then be an infinite regress of languages, each talking about the one below it. Observably, the grammar which describes the sentences of a language can be stated in sentences of the same language. This has obviously important effects, including the possibility of inserting metalanguage statements into the very sentences about which the metalanguage statement was speaking. At least one form of the complete grammar of the language is finite." [37]

[36] Cf. John Lehrberger, "Sublanguage Analysis", in Ralph Grishman and Richard Kittredge, *op. cit.*, p.20.

Moreover, Harris proposed for certain scientific texts a distinction between a science language component, accounting for the properties of objects and relations between them in a given realm of scientific inquiry and a meta-science component, which describes the relationship of the researcher / observer to his own methods and results[38]. A sentence may very well embody both components. That is what happens, for instance, anytime a clause of science sublanguage is embedded in a meta-science predicate. The proper analysis of such mixed sentences constitutes a prerequisite to the choice of the most adequate parsing strategy and dictionary construction for the automatic analysis of such texts. However abstract it may appear to be, the meta-science component can never abandon its roots in the empractic field (*empraktisches Umfeld*, in Karl Bühler's terminology), no matter how oblivious to that fact its everyday practice or the ambitions of its human operators be[39].

"There is no way to define or describe the language and its occurrences except in such statements said in that same language or in another natural language. Even if the grammar of a language is stated largely in symbols, those symbols will have to be defined ultimately in a natural language." [40]

Such claims deeply relate to what has become common knowledge within Quine's or Tarski's Semantics (search of truth of the propositions), Formal Logic and

---

[37] Cf. Zellig Harris, *Mathematical Structures of Language*, Robert E. Krieger Publishing Company, Huntington, New York, 1979, p. 17.

[38] This view was implemented, for example, by Naomi Sager in her analysis of the sublanguage of experimental pharmacology. "Syntactic Formatting of Scientific Information", *Proceedings of the 1972 Fall Joint Computer Conference*, *AFIPS Computer Conference,* Montvale, AFIPS Press, 1972, p. 791-800.

[39] Attention to this aspect of scientific inquiry constitutes one of the main traits of Michel Bitbol's important book *Physique et Philosophie de l'Esprit*, Paris, Champs-Flammarion, 2000.

[40] Cf. Zellig Harris, *A Theory of Language and Information: A Mathematical Approach*, Oxford, Clarendon Press, 1991, p. 274.

Metamathematics, and have been echoed in even more radical formulations in other linguistic traditions[41]:

"[…] por encima, y ocasionalmente por debajo, de los varios niveles de lenguajes formalizados que sucesivamente tomen como objeto de su habla fórmulas o términos del lenguaje "anterior" o "inferior", hay un lenguaje supremo (aunque nada más se presente en el prólogo del escrito) que es un lenguaje "natural" o no formalizado: *nuevamente* "natural" o no formalizado en el caso, al menos, de la Semántica, cuando el lenguaje de nivel "más bajo" al que se refería era ya un lenguaje "natural"."

Or, as Zellig Harris, puts it:

"The possibility of stating metalinguistic sentences within the grammar of the language makes the language describable within itself. And the fact that these contain metatoken sentences, as well as metatype, makes reference describable within the sentence itself. Otherwise, the description and interpretation of the sentence structure and of references in the sentence (pronouns, etc.) would have to be done in a separate metalanguage which would in turn have to be defined in a separate metalanguage of it, and so on without end." [42]

Harris views the metalanguage as a structurally distinguished sublanguage[43]. Its two main traits relevant for our purpose can be summarized as follows: it allows for clarification of very specific syntactic problems (e.g., certain derivations for tense or another one for reference from cross-reference) and it makes the whole of a natural language self-contained.

---

[41] Cf. Agustín García Calvo, "Tentativas para precisar la imprecisión del uso de los términos *significación, denotación* y *sentido, metalingüístico* y *abstracto, pragmático* y *modal*", in *Hablando de lo que habla*: *Estudios de lenguaje*, Zamora Editorial Lucina, 1990 [3ª ed.], p. 34.

[42] Cf. Zellig Harris, *op. cit.*, p. 202.

[43] Cf. Zellig Harris, "The background of transformational and metalanguage analysis", in Bruce Nevin (ed.), *The Legacy of Zellig Harris: Language and information into the 21st Century, Vol. 1: Philosophy of science, syntax, and semantics*, Amsterdam / Philadelphia, John Benjamins, 2002, p. 8.

### §5.4 Structuralist Metatheory of Structural Linguistics?

In a somewhat ironic way, Harris's approach is prone to be treated as an argument by a very particular function, namely, what is usually referred to as Metatheory of Science, which considers theories to be classes of model-theoretic structures[44]. Linguistics has not been alien to such treatment. Thus, as it has been already mentioned, Meier (2012) carries out a logical reconstruction of Leonard Bloomfield's theory.

"A set of **potential models** ($M_p$) fixes the general framework, in which an actual model of a theory is characterized. All entities that can be subsumed under the same conceptual framework of a given theory are members of the sets of the potential models of this theory. Sets of **partial potential models** ($M_{pp}$) represent the framework for the corroboration or refutation of the theory in question, they represent the framework of data, which shall corroborate or refute a theory. The concepts in $M_{pp}$ can be determined independently of T. Terms which are theoretical (and proper to T) in the potential models of the respective theory are cut out. Sets of models which do not only belong to the same conceptual framework, but also satisfy the laws of the same theory are called the sets of **actual models** ($M$) of a theory T. Local applications of a theory may overlap in space and time. The sets of **global constraints** ($GC$) are formal requirements that constrict the components of a model in dependence of other components of other models. Constraints express physical or real connections between different applications of a theory, i.e. the inner -theoretical relations. The sets of **global links** ($GL$) represent the intertheoretical connections between different theories."[45]

The idea of *specialization of scientific laws* plays a crucial role within this framework, as well as the criterion leading to it. This is perfectly illustrated by the following paragraph:

---

[44] The first endeavors in this direction can be traced back to Suppes (1957), who resorts to set-theoretic predicates in order to outline, in an axiomatic manner, the logical structure of a given empirical theory. Interestingly enough, Chomsky's 1955 *The logical structure of linguistic theory* partially overlaps in its goals with such enterprise. During the last decades, many case studies out of different branches of science have been carried out along these lines: see, for instance, Balzer, and Moulines, 1996, or Díez and Lorenzano, 2002.

[45] Cf. Thomas Meier, *op. cit.*, p. 12.

“When considering our reconstruction of Classical Particle Mechanics, the reader might already have asked himself where we have left such important laws of classical particle mechanics as Newton’s third law (the actio-reactio principle), the law of gravitation, or Hooke’s law. Our answer is: They all constitute different but interrelated theory-elements of classical particle mechanics. The whole array, in turn, constitutes what we might call “the theory-net of classical particle mechanics”. The same holds for other advanced theories of empirical science. In the case of simple equilibrium thermodynamics, besides the fundamental equation and the constraints and links (which, admittedly, provide much of the content of this theory), one would like to see Nernst’s “Third Principle of Thermodynamics”, Gay-Lussac’s law, and other more special laws. Many of these more special laws of the theory are, moreover, associated with particular constraints and, possibly, particular links, besides those already explicated when dealing with the “basic” theory-elements. In other words, the consideration of all these further requirements will end up in the reconstruction of a whole series of different theory-elements, which, however, have the same basic structure. Because of this similarity of structure, we can speak of a theory-net and not just an amorphous set of single, isolated theory-elements.”[46].

Meier’s reading of Leonard Bloomfield epistemology of linguistics along the lines of specialization laws certainly paves the way for a fruitful interpretation of Harris through that looking glass.

## §5.5 Operators

One of Harris’s most remarkable contributions lies in what has been called *Operator grammar*. The so-called *adjoint operators* represent a basic element of the theory. There is an important (and probably tight) relation between adjoint operators and right ideals. Richard Kittredge[47] correctly reminds us that the way in which certain sentential types (causal sentences, for instance) are adjoined to the right of narrow sublanguage sentences under special conjunctions suggests, as Harris himself first pinpointed, the algebraic structure of a right ideal in a ring,

---

[46] Cf. Wolfgang Balzer, Carlos Ulises Moulines and Joseph Sneed, *An Architectonic for Science*, Dordrecht, Reidel, 1987, p.168.

[47] Cf. Richard Kittredge, *Embedded sublanguages and natural language processing*, Proceedings of the 8th conference on Computational linguistics, 1980, p. 209.

There are, as we have seen, significant differences in lexicon and grammar between the narrow sublanguage portion of a given text and the adjoined sentences from the broader system.

An appropriate parser is expected to recognize the junctures between the broad and narrow portions, and to resort to the constraints present in the embedded sublanguage, while at the same time parsing the sentences of the loose matrix.

"… certain sublanguages have tightly structured "cores" which are embedded in a looser matrix whose lexical restrictions are closer to those found in the general language." [48]

The term 'loose matrix' is to be understood as synonym with the more conventional 'sparse matrix', as opposed to 'dense matrix'[49]. The former refers to matrices in which only non-zero elements are stored, whereas the latter names fully-fledge version, including zero entries. It goes without saying that a loose or sparse matrix can be created out of a dense matrix, by means of different formats[50].

*Sparsity* is defined as the fraction or the quotient which results from dividing the number of non-zero elements by the total number (m x n) of entries in the matrix. Thus, in a 4 x 4 matrix with 5 zero entries, sparsity equals 11 / 16. From a conceptual standpoint, sparsity is a trait of systems which are loosely coupled, in which, to just name a case scenario, only adjacent elements are connected, as opposed to each element

---

[48] Cf. Richard Kittredge, "Introduction" to Richard Kittredge (ed.), *Sublanguage: Studies of Language in Restricted Semantic Domains*, Berlin / New York, De Gruyter, 1982, p. 4. It is indeed of utmost interest what the same author claims elsewhere ("Variation and Homogeneity of Sublanguages", in *op. cit.*, p. 136):

"If the junctures between embedded sublanguage and matrix are identifiable, the strict sublanguage grammar rules […] can be limited to apply only on those clauses or clause fragments of the embedded sublanguage. When the parser scans matrix material, a more general parsing strategy and lexicon can be called upon. Such a dual processing strategy would make a processing system more powerful and less vulnerable to unfound words and structures, since as a rule these would occur in the loose matrix, which is less closed as a system."

[49] Although it might seem that Kittredge is concerned here with the qualitative aspect of the relationship between sublanguages rather than with solving large sparse systems of linear equations, the fact is that he also foresees some role for numerical linear algebra in the treatment of the whole issue.

[50] Formats essentially try to respond to two competing demands: efficient modification versus efficient access. Those derived from the former are the ones used to construct the matrices, whereas the latter facilitate matrix operations.

being linked to all others in a tight network susceptible of being represented by a dense matrix[51].

The Yale sparse matrix format is a good example of a loose matrix: it stores an initial sparse $m \times n$ matrix, M, by means of three one-dimensional arrays (labeled A, IA, and JA)[52].

Not only does the number of non-zero entries play a role, but also their distribution. Depending on them both, different data structures can be implemented which yield huge savings in memory when compared to the basic approach in which all the original entries are preserved and processed. There is a trade-off, though: access to the individual elements becomes much more complex and additional structures are required to recover the original matrix in a unambiguous manner.

It should be mentioned on passing that the product of a sparse matrix with itself always renders a sparse matrix. However, the number of zero entries may vary, so that the result might not be as optimal as the initial state.

Algorithms for sparse matrices are normally more complicated than their dense counterparts. For instance, when factorizing the matrix, attention must be paid to the ordering and fill-in issues, since any of them might destroy the sparsity of the factors, thereby rendering the use of sparse structures meaningless.

Technically speaking, Harris makes use of an *operator precedence grammar*, a type of context-free grammar with the following salient property: none of its productions can be empty in their right-hand side or present two adjacent non-terminals

---

[51] As a matter of fact, 'sparsity' has no less than four meanings, regarding, respectively: 1) Data sparsity, leading to the so-called "curse of dimensionality", 2) Probability sparsity, referred to a probability distribution over events, 3) Sparsity in the dual, which relates to the representation of predictors via kernel-based methods, and 4) Model sparsity, which arises from deleting all not needed zero-valued dimensions from the vector function responsible for the encoding and modelling of the phenomenon at stake. It is the fourth meaning that the focus lies on here.

[52] The three arrays can be spelled-out as follows: let NNZ denote the number of nonzero entries in M. Zero-based numbering will be used for the indexes.

The array A is of length NNZ; nonzero entries are arranged left-to-right top-to-bottom. The array IA has length $m + 1$. It contains the index in A of the first element in each row; the last element of the array corresponds to the total number of nonzero elements in M. IA $[i]$ stands for the index in array A of the first nonzero element of row $i$, while IA $[n]$ is NNZ: it follows then that the indices in A of nonzero elements of row $i$ of M are IA $[i]$ to IA $[i+1] - 1$. Finally, the vector JA is also of length NNZ, as it pertains to the column index in M of each element of A.

in it. In other words, precedence relations are to be defined between the terminals of the grammar. A parser is to be built up accordingly: one of the purest illustrations thereof is provided by Edsger W. Dijkstra's shunting yard algorithm[53].

The whole operator-argument system is then construed as a mathematical object warranting the stability of language structure.

## §6. Summary

The study of sublanguages has received some scholarly attention from its inception. Practical applications have not been alien to it either. However, little thought has been devoted to their very structure, that is, to the mathematics underlying them. In my view, it is precisely the reflection on such matter that should provide a major step to a better and deeper understanding of the nature of sublanguages and their embedding in a broader grammar. Even though, it might be argued, Noetherian rings have been covertly incorporated into some schools of linguistic thought via a somewhat *ad hoc* instantiation of the ascending chain condition, the fact remains that, other than among computational linguists and computer scientists with focus on natural language[54], no awareness of their importance is to be found in the field.

The future publication of a case study, pending funding approval, on the sublanguage of theoretical physics drawn from a short selection of texts from the classical period of the 20th century, responsible for the transition from relativity to quantum physics, will seek to deepen some of the issues addressed in the present paper.

---

[53] Cf. Edsger Wybe Dijkstra, "ALGOL-60 Translation", *ALGOL Bulletin*, Supplement 10, Amsterdam, 1961. The visual representation of the shunting yard algorithm starts on page 7 of the second paper included in that work, namely, "Making a Translator for ALGOL 60".

[54] Lambek syntactic calculus constitutes one of the most remarkable examples. For its potential regarding the description of natural language, see, for instance, Sean A. Fulop, "Learnability of type-logical grammars", *Electronic Notes in Theoretical Computer Science*, 53, 2001, 14 pages.

Finally, it has to be noted that the connection between the notion 'minimal ideal' and that of 'minimalism' in its current programmatic sense in the theory of language and linguistics, though far for obvious, should be worth exploring.

## 7. Final Remarks

One might ask, in the light of the Noetherian definition of one-sided ideals, where the difference with Polish (and reverse Polish) notation lies. At first sight, it might seem like the contrast between right and left ideals just mirrors that between prefix and postfix notation. Both share the property of their operators presenting a fixed arity, which leads to unambiguous parsing. Upon closer inspection, they are substantially different, though. The key criterion is provided by the following statement in Noether's seminal paper: "Jedes Element r definiert zugleich zwei Operatoren: die Operatoren rx und xr". Two operators are simultaneously defined, not just one, as it would be the case in any form of the Polish notation. In fact, Polish notation makes the relationship more obscure than desired: the difference mentioned by Noether corresponds to that between $r\ a\ b\ {+}{\times}$ and $a\ b + r\ \times$.

With regard to left and right modules, the distinction is applied over non-commutative rings, in order to specify the side where a scalar $s$ (or $t$) appears in the scalar multiplication. In the case of a bimodule, that is, a mathematical object being simultaneously a left and right module, two different scalar multiplication operations are implied.

Schematically:

| **Left module** | **Right module** |
|---|---|
| $s(\mathbf{x} + \mathbf{y}) = s\mathbf{x} + s\mathbf{y}$<br>$(s_1 + s_2)\mathbf{x} = s_1\mathbf{x} + s_2\mathbf{x}$<br>$s(t\mathbf{x}) = (s\ t)\mathbf{x}$ | $(\mathbf{x} + \mathbf{y})t = \mathbf{x}t + \mathbf{y}t$<br>$\mathbf{x}(t_1 + t_2) = \mathbf{x}t_1 + \mathbf{x}t_2$<br>$(\mathbf{x}s)t = \mathbf{x}(s\ t)$ |

It is crucial for our purposes to emphasize that the distinction is not purely syntactical, since it implies two different associativity rules linking multiplication in a module with multiplication in a ring. Obviously, in the commutative case, where R =

$R^{op}$, the distinction is meaningless. But, as we know[55], there are rings in which $R \neq R^{op}$.. The rings sublanguages allegedly relate to seem to belong to the second group.

In category theory the usage of “left” and “right” may seem to bear some algebraic content but refers rather to left and right sides of morphisms. Adjoint functors are a perfect instantiation of it.

A last word is due: the linguistic status of the mathematical objects treated here (or the objects themselves, for that matter) ought to be interpreted pretty much as the non-numerical existence of vectors previous to and independent from any basis choice, as could be inferred from the philosophy underlying some of Edward Frenkel’s recent conferences for a wider public.

---

[55] Remember the case of matrices in section §.5. Else, consider triangular rings of the form

$$A = \begin{pmatrix} R & M \\ 0 & S \end{pmatrix}$$

(in which R and S are rings and M is an R-S-bimodule), such as the finite ring

$$\begin{pmatrix} Z/4Z & Z/2Z \\ 0 & Z/2Z \end{pmatrix}$$

which has 11 left ideals and 12 right ideals. The left ideals in the case of this ring type are isomorphic to $U \oplus J$, where J is a left ideal of S, and U an R-submodule of $R \oplus M$ with $MJ \subseteq U$. There are plenty of similar examples, following from the existence of many ring theoretic notions which are not left-right symmetric. For a wide range of them, see Tsit Yuen Lam, *Lectures on Modules and Rings*, Graduate Texts in Mathematics 189, New York, Springer, 1999, and *A First Course in Noncommutative Rings*, Graduate Texts in Mathematics 131, New York, Springer, 2001.
Torsion elements of the Brauer group with order other than 2 make up another pattern furnishing a further gamut of examples.